\documentclass[pdflatex,sn-mathphys-num]{sn-jnl}

\usepackage{graphicx}%
\usepackage{multirow}%
\usepackage{amsmath,amssymb,amsfonts}%
\usepackage{amsthm}%
\usepackage{mathrsfs}%
\usepackage[title]{appendix}%
\usepackage{xcolor}%
\usepackage{textcomp}%
\usepackage{manyfoot}%
\usepackage{booktabs}%
\usepackage{algorithm}%
\usepackage{algorithmicx}%
\usepackage{algpseudocode}%
\usepackage{listings}%
\usepackage{natbib}
\usepackage{caption}
\theoremstyle{thmstyleone}%

\theoremstyle{thmstyletwo}%

\theoremstyle{thmstylethree}%

\begin{document}

\title[Article Title]{What to randomize matters: a controlled study of volume, axis and representation in low-data molecular generation}

\author*[1]{\fnm{Neerav} \sur{Kaushal}}\email{kaushal@mtu.edu}
\author[1]{\fnm{Alisa} \sur{Teige}}\email{ateige@mtu.edu}
\equalcont{These authors contributed equally to this work.}
\author[1]{\fnm{Ajay MNV} \sur{Penmatsa}}\email{mpenmats@mtu.edu}
\equalcont{These authors contributed equally to this work.}
\affil*[1]{\orgdiv{Department of Data Science}, \orgname{Michigan Technological University}, \orgaddress{\street{1400 Townsend Dr}, \city{Houghton}, \postcode{49931}, \state{MI}, \country{USA}}}

\abstract{Chemical language models can struggle to generate new molecules when trained on small datasets, where a model may reproduce training examples rather than learn a useful generative distribution. Randomized SMILES augmentation varies atom traversal, whereas Sequential Attachment-based Fragment Embedding (SAFE) also allows the order of fragment blocks to vary. How these distinct sources of representational variation affect low-data molecular generation has not been clearly studied. Here, using 1,000 seed molecules to train an autoregressive sequence model, we compare canonical SAFE with randomized SAFE at increasing corpus sizes, isolate atom-order and fragment-order axes of SAFE randomization, and compare fully randomized SAFE with randomized SMILES at molecule-by-molecule matched training corpora. Three findings emerge. First, randomization of SAFE alone raises the genuine yield of novel molecules more than 450-fold, turning a model that merely reproduces its training set into a practical molecular generator. Second, the two axes of randomization act jointly rather than additively: atom order alone reaches a tenth, and fragment order alone a seventh, of the genuine yield obtained when both vary, which is four times their combined contribution. Third, at matched corpus size and compute, randomized SAFE delivers 2.8 times the genuine yield of randomized SMILES. Welch tests with Games--Howell post-hoc comparisons resolve every difference on the primary outcomes. In the low-data regime, thus, \textit{what} one randomizes rather than only \textit{how much} determines whether augmentation produces new molecules or merely new spellings of the training set.

\textbf{Scientific Contribution:} This study compares string augmentation strategies in the low-data regime, where randomization has been reported to offer limited benefit at benchmark scale, using matched corpora and compute to separate SAFE's two randomization axes and to compare randomized SAFE with randomized SMILES. The main findings are that randomization turns a memorizing model into a practical generator, that the two axes of randomization act jointly rather than additively, and that randomized SAFE exceeds randomized SMILES in performance at matched corpus size. This work contributes the first augmentation axis-resolved analysis showing that augmentation design, and not augmentation volume alone, determines molecular-generation outcomes in data-poor settings.}


\maketitle

\section{Introduction}\label{sec:intro}

In the last decade, deep learning-based generative models have become central to de novo molecular design~\cite{gomez-bombarelli_automatic_2018, sanchez-lengeling_inverse_2018, oliveira_deep_2025, lee_genmol_2025, cheng_molecular_2021, tang_survey_2024, sousa_generative_2021} and among the most widely used ones are autoregressive language models trained on string representations of molecules~\cite{segler_generating_2018, moret_leveraging_2023, ozcelik_chemical_2024, bagal_molgpt_2022}. These benchmark models are mostly trained on vast swaths of data ranging from millions to billions. Yet the settings in which generation would be most valuable are often the most data-poor. A medicinal chemistry campaign may have only tens to a few hundred molecules of interest around a chemotype or disease target. Learning a useful generative distribution from the low-data regime of around 1,000 molecules or fewer is therefore a recurring, practical, and important challenge in the field of molecular generation and drug discovery.

Most molecular language models operate on Simplified Molecular Input Line Entry System (SMILES)~\cite{weininger_smiles_1988}, a string built by traversing a molecule atom by atom. A single molecule can be represented by many valid SMILES strings and this redundancy has been exploited as data augmentation~\cite{bjerrum_smiles_2017, arus-pous_smiles-based_2020, skinnider_chemical_2021, ozcelik_chemical_2024}. Enumerating multiple non-canonical or randomized SMILES per molecule was shown by~\citet{arus-pous_smiles-based_2020} to improve the validity, novelty, and distributional coverage of SMILES-based generative models. SMILES randomization exposes the model to many spellings of the same molecule, discouraging memorization of a single canonical form and increasing learning robustness. More recent work extends this idea with SMILES-level operations such as token deletion, atom masking, and bioisosteric substitution for low-data generation~\cite{brinkmann_going_nodate, grisoni_chemical_2023}. All of these operate on the SMILES grammar, whose only source of randomization is the order in which atoms are traversed.

\begin{figure}[h]
\centering
\includegraphics[width=0.6\textwidth]{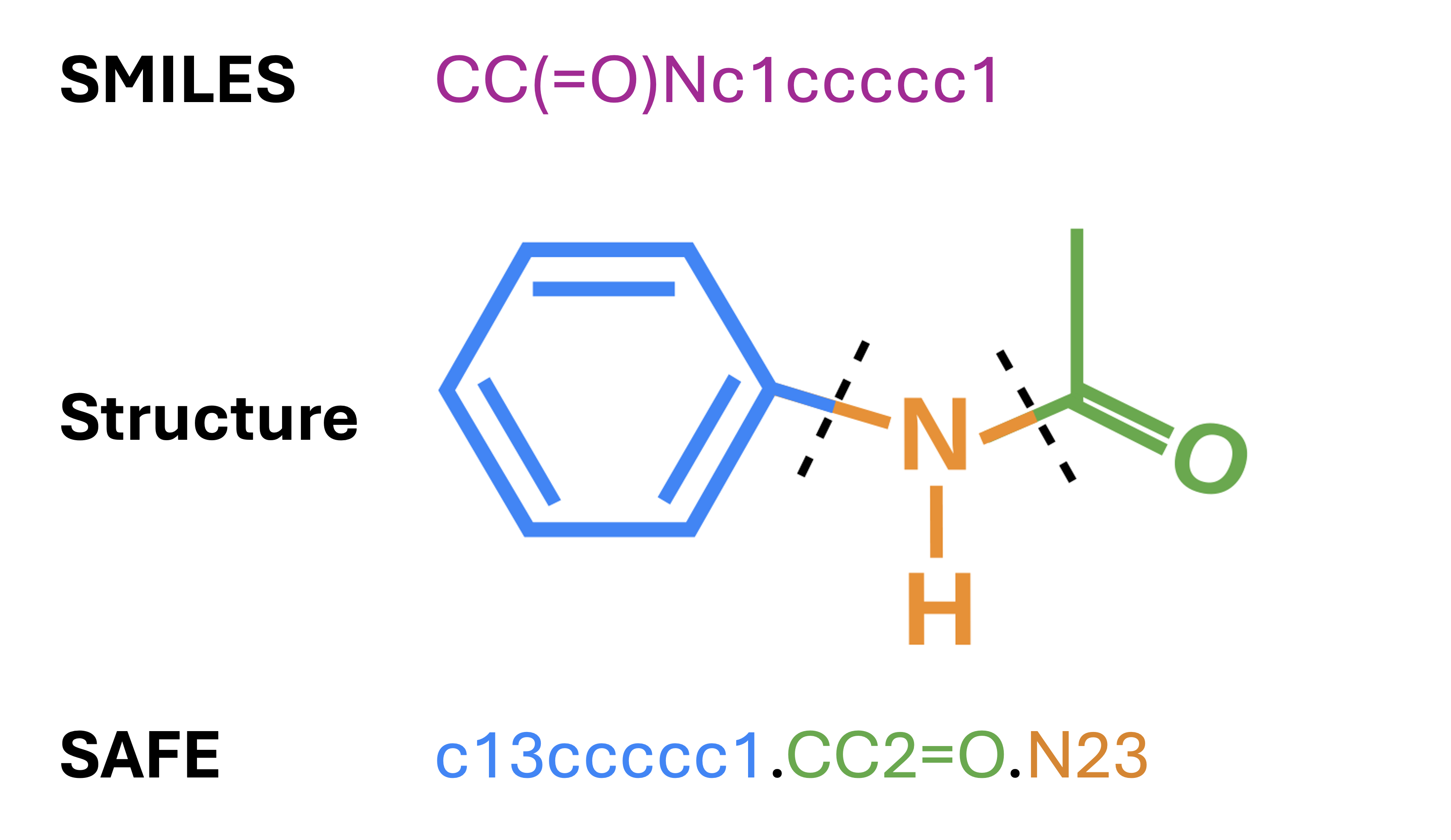}
\caption{\textbf{The SAFE representation}. The same molecule (acetanilide) is shown three ways: as a SMILES string (top), a 2D structure (middle), and a SAFE string (bottom). SMILES encodes a molecule by an atom-by-atom traversal. SAFE instead decomposes the molecule into fragments, here cleaved by the BRICS algorithm into a phenyl ring (blue), the central nitrogen (orange), and an acetyl group (green), with dashed lines marking the cleaved bonds. Each fragment gets written as a SMILES-like block. The blocks are concatenated with a {\Huge$\mathbf{.}$} separator, and the bonds that originally joined the fragments are preserved as matching ring-bond attachment digits shared across blocks (e.g.\ the shared \texttt{2} and \texttt{3}). Colors link each fragment to its corresponding block.}
\label{fig:safe}
\end{figure}

Sequential Attachment-based Fragment Embedding (SAFE)~\cite{noutahi_gotta_2024} representation is an alternative, information-equivalent linearization organized by molecular fragments rather than atoms. It preserves fragment composition, connectivity, and scaffold context. A molecule is first decomposed into fragments by a slicing scheme---obtained by BRICS~\cite{degen_art_2008} decomposition in our work---which cuts retrosynthetically accessible bonds, resulting in each fragment being written as a SMILES-like block. Fragments are concatenated into a single string, separated by {\Huge$\mathbf{.}$} (dot), and the bonds that originally joined them are preserved as matching ring-bond digits shared across blocks, which act as attachment points. For example, acetanilide becomes \texttt{c13ccccc1.CC2=O.N23}, constituting a phenyl block, an acetyl block, and a nitrogen block, with the shared digits \texttt{2} and \texttt{3} encoding the \texttt{N}---\texttt{C(=O)C} and \texttt{N}---\texttt{phenyl} bonds respectively. This is shown in Figure~\ref{fig:safe}.

\begin{figure}[h]
\centering
\includegraphics[width=0.95\textwidth]{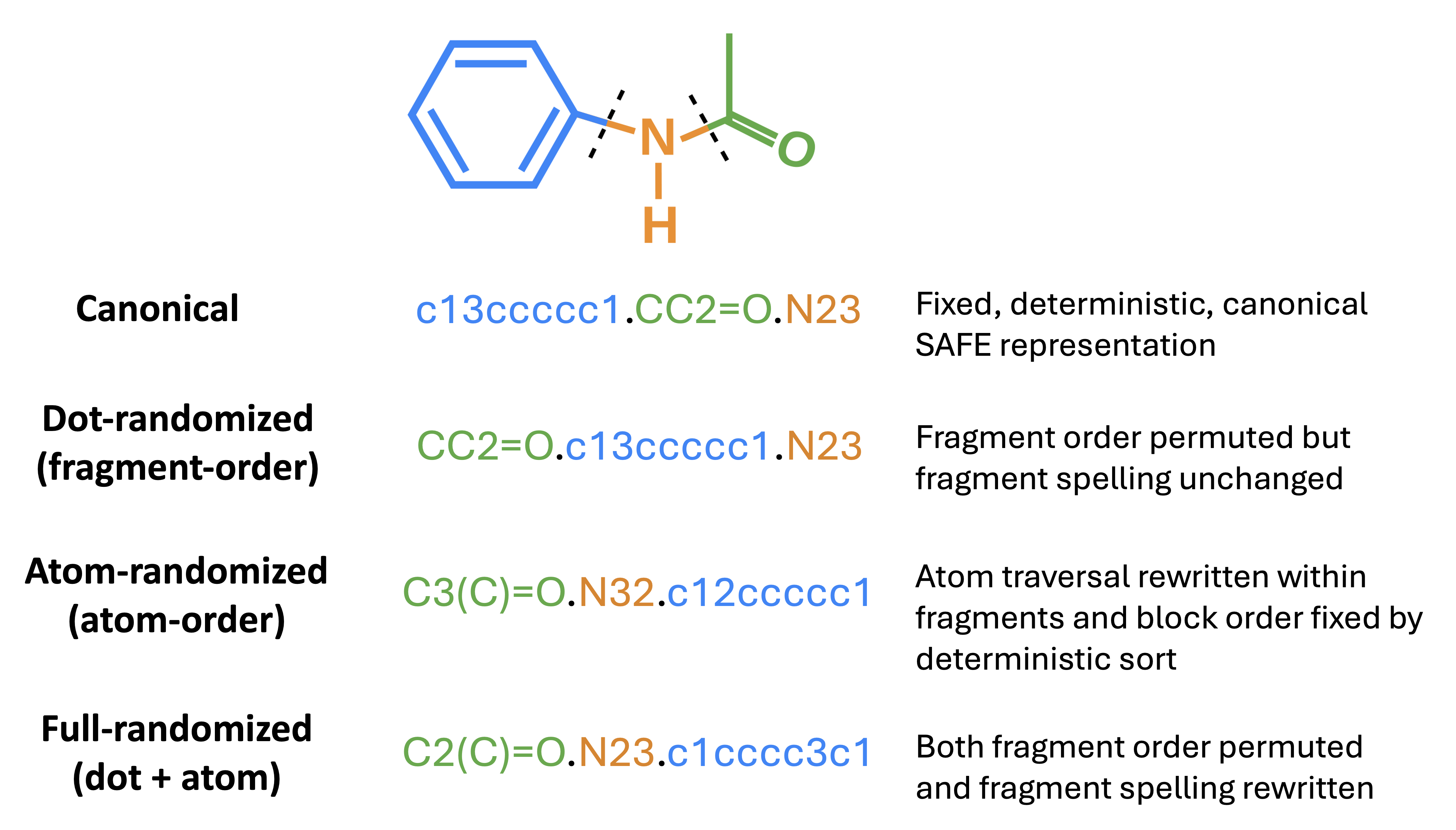}
\caption{\textbf{The axes of randomization of a SAFE string}. All four SAFE strings encode the same molecule, acetanilide (top) and decode back to it. Colors mark the three BRICS fragments (phenyl:blue, acetyl:green,
and nitrogen:orange). \emph{Canonical}: the single deterministic SAFE string. \emph{Dot-randomized (fragment-order)}: the fragment blocks are permuted while each
block's internal spelling is unchanged, meaning the atom traversal within a fragment follows the same order as in the canonical representation. \emph{Atom-randomized (atom-order)}: the atom traversal within each fragment is rewritten, and the resulting blocks are then arranged in deterministic sorted order rather than a random one, so that randomness enters only through the spelling of each fragment and not through their sequence. \emph{Full-randomized (dot\,+\,atom)}: both axes vary jointly, the default for SAFE randomizer. Because a molecule's fragments can be reordered and each fragment respelled independently, SAFE exposes two orthogonal randomization axes. SMILES randomization varies only the atom-order axis, making the fragment-order axis unique to SAFE.}
\label{fig:safe_axes}
\end{figure}

Like SMILES, a SAFE string is a valid, invertible encoding that a language model can generate token by token and that can decode back to a unique molecule.  Unlike SMILES though, its fragment structure is explicit in the string. This makes SAFE natural for fragment-constrained generation tasks as shown by~\citet{zhu_conditional_2025}, and it introduces a second source of string variation that SMILES does not possess: the order of the fragment blocks themselves, which is central to this work. Critically, randomizing a SAFE string can be done by varying two orthogonal axes: the order of fragments and the atom ordering within the fragments, whereas SMILES randomization varies only the atom ordering. This is shown in Figure~\ref{fig:safe_axes} and discussed in detail in Sec~\ref{subsec:data}.

Whether augmenting data by SAFE randomization benefits molecular generation at all, and if so, whether that benefit arises from the fragment-order axis, the atom-order axis, or their combination, is unestablished. Indeed, at benchmark scale (MOSES~\cite{polykovskiy_molecular_2020}, $10^4$-$10^6$ molecules), \citet{mesbahi_safe_2024} reported that SAFE randomization yields limited benefit, attributing what benefit there is largely to extended training and reduced overfitting rather than to improved generalization. These observations leave three questions open, each bearing directly on when and why representation-level randomization helps:
\begin{enumerate}
    \item Does SAFE randomization help in a strictly low-data regime ($\sim 10^3$ molecules), as SMILES randomization does, or is its benefit as limited there as at benchmark scale?
    \item SAFE randomization couples two axes: fragment order and atom order. Which, if either, or both, is responsible for any benefit?
    \item If there is indeed a benefit, is it due to the SAFE representation itself or to augmentation in general? Since only SAFE has the fragment-order axis, a
    controlled comparison against baseline SMILES randomization at matched data and
    compute can tell the two apart. 
\end{enumerate}

To our knowledge, none of these questions has been addressed directly. We approach them with a controlled study built around 1,000 seed molecules and a small decoder-only transformer (OPT) trained from scratch. Across all comparisons we hold the architecture, tokenization scheme, and optimizer-step budget fixed. For the axis ablation and the SAFE--SMILES comparison we
additionally match the molecular composition of the training corpora molecule by molecule, so that those arms differ only in the randomization applied. Each arm is trained five times from independent initializations and evaluated on validity, uniqueness, novelty, yield, genuine yield, seed recovery, internal diversity, and physicochemical property distributions. Three results follow:

\begin{enumerate}
\item In this low-data setting, SAFE randomization changes the character of the model's output rather than merely improving it. Genuine yield rises from 15 to 7,079 novel molecules per 10,000 samples, so a model that largely reproduces its training set becomes a working molecular generator. This contrasts with the limited benefit reported for SAFE randomization at benchmark scale~\cite{mesbahi_safe_2024}.

\item Neither randomization axis alone approaches their combination. At matched composition, varying both axes of randomization together reaches 6,977 genuine-yield molecules per 10,000 samples, against 712 for atom-order and 1,026 for fragment-order randomization. The two axes thus interact, rather than add.

\item Under identical training and evaluation conditions, fully randomized SAFE produces 7,067 genuine-yield molecules per 10,000 samples against 2,562 for randomized SMILES, a gap that widens relative to the plain novelty count because SMILES randomization diversifies spellings more than the molecules.

\end{enumerate}

Taken together, these comparisons indicate that an augmentation scheme is better characterized by the transformations it introduces than by the number of strings it produces. In other words, in the low-data regime, what one randomizes matters. The fragment-order axis afforded by SAFE's fragment grammar, and absent from SMILES, appears to be what converts string augmentation into molecular novelty.

\section{Methods}\label{sec:methods}

\subsection{Data and Input}\label{subsec:data}
We study a deliberately small training set to probe the low-data regime. 1,000 molecules were sampled uniformly at random from the GuacaMol~\cite{brown_guacamol_2019} training collection with a disjoint set of 200 molecules held out for validation. All molecules were standardized and canonicalized with RDKit~\cite{greg_landrum_rdkit_2026}. Each molecule was encoded in two ways: as its RDKit canonical SMILES, and as a SAFE string produced with the reference implementation~\cite{noutahi_gotta_2024} using BRICS fragmentation, as described in Section~\ref{sec:intro} and illustrated in Figure~\ref{fig:safe}. The framework is agnostic to the fragmentation scheme (RECAP, Hussain--Rea, MMPA, or other slicing algorithms would serve equally), so our choice of using BRICS is not fundamental. It does, however, set the fragment granularity, and hence the number of fragments per molecule, which bounds the diversity achievable by fragment-order randomization. Given the two SAFE randomization axes, fragment order and atom order, we generate four SAFE variants (see Figure~\ref{fig:safe_axes}) and one SMILES variant:

\begin{itemize}
    \item \textbf{Canonical SAFE}: The single canonical SAFE string. This yields exactly one string per molecule and applies no augmentation, serving as the unaugmented baseline against which the three randomized schemes are compared.

    \item \textbf{Dot-randomized SAFE (Fragment-order)}: The canonical SAFE string with its fragment blocks randomly permuted, leaving the atom ordering (and thus spelling) within each fragment canonical. Fragment reordering is always valid in SAFE because attachment is defined by matched ring-bond digits rather than position.

    \item \textbf{Atom-randomized SAFE (atom-order)}: The randomized atom ordering within each fragment block, followed by a deterministic sorting of the fragment blocks, which removes fragment order as an independent source of variation and isolates atom-order variation.
    
    \item \textbf{Full-randomized SAFE (atom- and fragment-order randomization)}: Both axes varied jointly, as in the original SAFE reference.

    \item \textbf{Randomized SMILES}: non-canonical or randomized SMILES generated by a random atom permutation; atoms are renumbered by a random permutation and the molecule is written non-canonically.
    
\end{itemize}

For the atom-randomized axis, we isolate within-fragment atom variation from fragment-order variation. We do so by randomizing at the whole-molecule level so that the shared ring-bond attachment digits that bond the fragments remain mutually consistent and then reordering the resulting fragment blocks by sorting, rather than respelling each fragment in place. Because a fragment's attachment points are encoded as ring-bond digits shared with its neighbors, respelling a fragment in isolation would reassign or break those digits and can yield invalid or wrongly connected strings. Sorting the blocks instead removes fragment order as an independent source of variation while keeping the attachment intact, leaving only the atom ordering within each fragment free to vary. Consequently, although an individual atom-randomized string can resemble a full-randomized string, across the corpus the fragment order is fixed by the sort rather than independently permuted, so unlike the full-randomized axis, the atom-randomized axis contains no free fragment-order randomization.

For every scheme we request up to $N$ distinct variants per molecule, seeded reproducibly, deduplicated, and always including the canonical encoding. We define an \textit{arm} as a single training configuration: a choice of string representation together with a randomization scheme. For example, SAFE with \texttt{dot}, SAFE with \texttt{atom}, SAFE with \texttt{full}, or randomized SMILES. Applied to the 1,000 seed molecules, each arm produces a corpus of training strings. A naive comparison that simply augments each arm to the same total string count confounds two effects: how many strings an arm contributes and which molecules those strings cover. Because arms differ in how many unique variants they can emit per molecule, equal totals can still leave the per-molecule composition unbalanced. For example, under the fragment-order axis, a small molecule with three SAFE fragments can yield at most $3! = 6$ distinct strings, whereas a larger molecule with six fragments can yield up to $6! = 720$; an arm that varies atom order has no such cap, so a fixed total would be filled mostly by larger molecules in one arm and spread more evenly across molecules in another. Thus, to isolate the effect of the randomization scheme from that of corpus composition, we match the arms molecule by molecule. For the set of arms $\mathcal{A}$ under comparison and $v_i^{(a)}$ being the number of unique/distinct strings available for molecule $i$ under arm $a \in \mathcal{A}$, we set the shared per-molecule
multiplicity
\begin{equation}
  k_i \;=\; \min_{a \in \mathcal{A}} v_i^{(a)},
  \label{eq:kmin}
\end{equation}
and draw exactly $k_i$ strings for molecule $i$ from every arm. The resulting corpora then share an identical composition wherein each arm contributes the same number of strings for the same molecule, and the matched total is
\begin{equation}
  N_{\mathrm{match}} \;=\; \sum_{i=1}^{1000} k_i ,
  \label{eq:nmatch}
\end{equation}
identical across all arms in $\mathcal{A}$. Any remaining performance difference between arms is therefore attributable to \emph{which} strings are drawn, not \emph{how many} or \emph{for which} molecules. We construct three corpus families, each targeting a distinct question: how much augmentation helps, which randomization axis drives the gain, if any, and whether the representation itself matters. These are:

\begin{enumerate}
  
  \item \textbf{Augmentation sweep.} Three SAFE arms, $\mathcal{A}=\texttt{safe-1k}, \texttt{safe-25k}, \texttt{safe-50k}$, all using the \texttt{full} randomization scheme which is the default in SAFE~\cite{noutahi_gotta_2024}, and trained on corpora of $1$K, $25$K, and $50$K strings, respectively. The arms deliberately differ in corpus size and are therefore not composition-matched (Eqs.~\eqref{eq:kmin} and~\eqref{eq:nmatch} do not apply); this family quantifies the data-efficiency gain from SAFE augmentation alone.
  
  \item \textbf{Axis ablation.} The three SAFE randomization schemes, $\mathcal{A}=\{\texttt{safe-dot},\texttt{safe-atom},\texttt{safe-full}\}$), matched to a common total of $N_{\mathrm{match}}=32{,}152$ strings per arm. This family isolates the contribution of each randomization axis of SAFE.
  
  \item \textbf{Representation comparison.} Randomized SAFE (with $\texttt{full}$) versus randomized SMILES, $\mathcal{A}=\{\texttt{safe-rand},\texttt{smiles-rand}\}$, matched to $N_{\mathrm{match}}=49{,}209$ strings per arm. This family isolates the effect of the string representation at fixed composition.
\end{enumerate}

\begin{figure}[h]
\centering
\includegraphics[width=\textwidth]{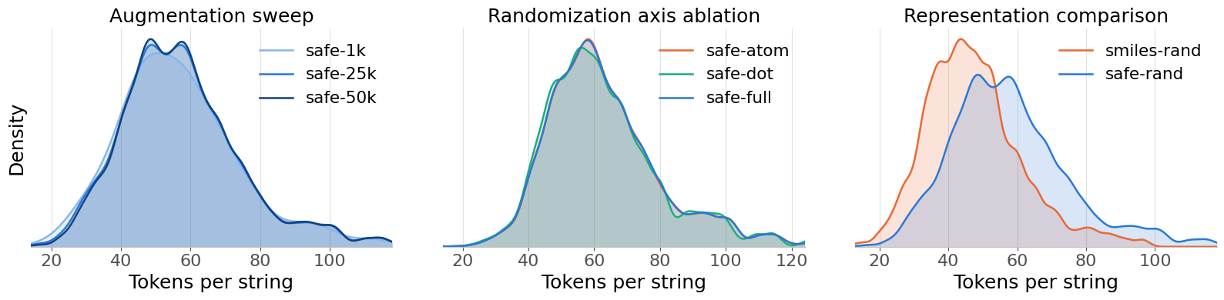}
\caption{\textbf{Token-length distributions of the training corpora}. Kernel density estimates of tokens per training string, excluding special tokens, with each arm under its own tokenizer. Distributions are closely matched within the augmentation sweep (left) and the axis ablation (middle), but SAFE strings are longer than randomized SMILES (right). See Table~\ref{tab:token-lengths} for summary statistics.}
\label{fig:token-lengths}
\end{figure}

Token-length distributions of all eight training corpora are summarized in Table~\ref{tab:token-lengths} and shown in Figure~\ref{fig:token-lengths}. Within each family the distributions are closely matched, as expected from the composition matching. The small shortfall of \texttt{safe-dot} relative to the other two axis arms reflects its preservation of canonical fragment spelling. Across representations the distributions differ because SAFE strings are longer than the corresponding randomized SMILES.

\begin{table}[!htbp]
\caption{\textbf{Summary statistics of token-length distributions of the training corpora}. Lengths are counts of tokenizer tokens per training string, excluding special tokens. SAFE and SMILES arms use their respective tokenizers.}
\label{tab:token-lengths}
\small
\begin{tabular*}{0.95\textwidth}{@{\extracolsep{\fill}}llrrcrrr@{}}
\toprule
Family & Arm & Strings & Median & Mean & SD & Max \\
\midrule
Augmentation sweep & \texttt{safe-1k} & 1,000 & 55 &56.8 & 17.8 & 132 \\
Augmentation sweep & \texttt{safe-25k} & 24,972 & 56 & 57.5 & 17.7 & 132 \\
Augmentation sweep & \texttt{safe-50k} & 49,450 & 56 & 57.8 & 17.6 & 134 \\
\midrule
Randomization axis ablation & \texttt{safe-atom} & 32,152 & 60  & 63.2 & 17.4 & 134 \\
Randomization axis ablation & \texttt{safe-dot} & 32,152 & 60  & 62.6 & 17.4 & 132 \\
Randomization axis ablation & \texttt{safe-full} & 32,152 & 60  & 63.2 & 17.4 & 134 \\
\midrule
Representation comparison & \texttt{smiles-rand} & 49,209 & 46  & 47.4 & 14.0 & 98 \\
Representation comparison & \texttt{safe-rand} & 49,209 & 56  & 57.9 & 17.5 & 134 \\
\botrule
\end{tabular*}
\end{table}

\subsection{Model, tokenizer, and matched-compute training} \label{subsec:mtmct}

\textbf{Tokenizer}: We use an atom-level tokenizer for the entire study. A regular-expression pre-tokenizer splits each string into chemically atomic units (bracket atoms, two-letter elements, ring-closure digits, bonds, branches, and the fragment separator), and the tokenizer's decoder reconstructs strings exactly. A separate vocabulary is fitted for each representation on its randomized corpus (51 tokens for SAFE and 38 for SMILES) reflecting that the vocabulary is a property of the representation. The tokenization scheme is held identical across all arms and we verified that zero unknown tokens occur in the corpora.

\textbf{Architecture}: We used a transformer-based~\cite{vaswani_attention_2017}, decoder-only, Open Pre-trained Transformer (OPT) Language Model~\cite{zhang_opt_2022} with identical hyperparameters across every arm. The model had a hidden size of 384, 5 layers, 8 attention heads, feed-forward dimension of 1,536, and a maximum sequence length of 140 tokens (chosen to exceed the longest tokenized string in every corpus), resulting in 8.95M parameters. Because we are addressing unconditional autoregressive generation, the models were trained with a causal language-modeling objective and neither masked language modeling nor sequence-to-sequence were used. All models share one architecture, hyperparameters, and parameter count, and are trained from scratch. They differ only in their training corpus: each corpus-model pair is an \textit{arm}, and the arms are grouped into the three
efforts (families) described above.

\textbf{Matched compute}: Every model is trained for an identical number of optimizer steps (20,000, with a batch size of 64, learning rate $5 \times 10^{-4}$, weight decay 0.01, and warmup ratio 0.01), so that arms are compared at equal compute. Because the augmented corpora are larger than the canonical one, the canonical arm makes correspondingly more passes over its data. This asymmetry is intrinsic to the low-data regime under study and is not a confound, since it is the direct consequence of having fewer distinct training strings. Each arm is trained five times with independent random seeds, which control
both weight initialization and the order in which training strings are presented, and the training corpus is held fixed across the five runs, so the five models of an arm differ only in their optimization trajectory. Validation loss on the 200 held-out molecules was monitored but not used for checkpoint selection and all reported results use the final (step-20,000) checkpoint. The training used HuggingFace~\cite{wolf_huggingfaces_2020} Transformers with
Accelerate~\cite{gugger_accelerate_2022} and was performed on NVIDIA RTX Pro Blackwell GPUs (48~GB), one training run per GPU.

\subsection{Generation and evaluation}\label{subsec:eval}
Decoding a generated SAFE string can result in an interesting scenario. Attachment is encoded by ring-bond digits shared between fragment blocks, so if the model emits a digit in one block without its partner elsewhere, the dangling bond is dropped and the string decodes to several disconnected, yet valid molecules. For example, the valid SAFE string \texttt{C3(C)=O.N32.c12ccccc1} decodes to acetanilide (SMILES: \texttt{CC(=O)Nc1ccccc1}) but omitting the final attachment digit \texttt{2} on the phenyl block (\texttt{C3(C)=O.N32.c1ccccc1}) leaves the digit \texttt{2} on the nitrogen unpaired, and the string decodes to two disconnected molecules, acetamide and benzene (SMILES: \texttt{CC(N)=O.c1ccccc1}). We therefore define the generative system as sampling followed by selection of the largest connected component whenever a sample decodes to more than one. This is a strict, deliberate choice we adopt. It makes sure that each parseable string contributes exactly one molecule, as in standard salt stripping, which reflects what such a model would deliver in practice. The alternative, counting every component of a multi-component decode as a separate generation, would raise yield for the SAFE arms while leaving the SMILES arm unchanged, since a single SMILES decodes to one component by construction. Our choice therefore works against the differences we report rather than producing them. The same pipeline is applied to every arm in all three families.


From each trained model we generate three independent sets of 10,000 strings (multinomial sampling, temperature 0.9, no top-$k$ or nucleus truncation). Every quantity is computed on each set separately and averaged within a training run, so each arm yields five run-level values, which we report as the mean $\pm$ standard deviation. Each parseable sample is first reduced to its largest connected component, as described above. For every arm we then report validity, uniqueness and novelty, together with internal diversity and Bemis--Murcko~\cite{bemis_properties_1996} scaffold diversity, both computed over the novel molecules. These quantities are conditional on one another (uniqueness is computed over valid molecules, and novelty over unique ones), so we also report \emph{yield}, the number of novel, unique, valid molecules obtained per 10,000 samples, which composes them into a single end-to-end count ($\texttt{yield} = N \times \texttt{validity} \times \texttt{uniqueness} \times \texttt{novelty}$). Yield is the primary outcome and it makes arms directly comparable without trading one component off against another, and a model that produces many valid strings but few distinct new molecules scores low. Two further measures address whether the novel molecules are genuinely new. First, we define \emph{Genuine yield} as the number of novel molecules whose nearest neighbor among the seed (train set) molecules has a Tanimoto similarity below 0.7 (Morgan fingerprints with radius 2 and 2048 bits). Novelty alone counts any molecule absent from the training and validation set including near-copies that differ from a seed by a single atom or bond, so genuine yield separates structurally new molecules from minor edits of the training data. The threshold of 0.7 is a structural near-duplicate cutoff rather than an activity-relevance criterion as single-atom and single-bond edits typically leave Morgan similarity above this value. Reported activity-relevant similarity values are fingerprint-dependent and lower for circular
fingerprints~\cite{martin_structurally_2002,jasial_activity-relevant_2016}, so 0.7 is a permissive cutoff and a stricter one would discard a larger share of the SMILES arm's output than of SAFE's and would therefore widen rather than narrow the differences we report in our results. Second, \emph{Seed recovery} is the fraction of the 1,000 seed molecules that the model reproduces exactly among its valid generated molecules. It directly measures memorization, since a high value means that much of the output reproduces the training set. These metrics are described in detail in Table~\ref{tab:metrics}. Novelty and genuine yield use different reference sets by design. Novelty is measured against the training and validation molecules, so that a molecule appearing anywhere in the study is not counted as new, while nearest-neighbor similarity is measured against the training molecules alone, since only those
could have been memorized. The 200 validation molecules alter novelty by less than 0.001 in every arm. No test set is required in this study. We do not evaluate predictive performance or select among models on held-out data. Every arm is trained for the same fixed number of optimizer steps and evaluated at its final checkpoint, and all metrics are computed on generated molecules relative to the training (and validation) molecules.

Differences among the corpus-size arms and among the three randomization axes are assessed by Welch's ANOVA with Games--Howell post-hoc comparisons, while the SAFE-versus-SMILES contrast is assessed by Welch's $t$-test. Welch’s tests are used unconditionally rather than selected on a preliminary variance test, because the arms differ markedly in variance: the ratio of the largest to the smallest per-arm standard deviation reaches 97 in the augmentation sweep. The unit of replication is the training run ($n=5$ independently trained models per arm), and we report mean differences with 95\% confidence intervals alongside $p$-values. Yield and genuine yield are the primary outcomes. All other metrics are secondary and not corrected for multiple comparisons, and internal diversity is reported descriptively. Full test results are given in Supplementary Tables S1–S5.

\begin{table}[!htbp]
\centering
\caption{\textbf{Evaluation metrics}. For one generated set, $n=10{,}000$ strings are sampled and each is decoded to a molecule, keeping its largest connected component. $\mathcal{V}$ is the resulting valid molecules, $\mathcal{U}$ the distinct molecules in $\mathcal{V}$ (compared as canonical SMILES), $\mathcal{T}$ the 1,000 training molecules, $\mathcal{R}$ the reference set of $\mathcal{T}$ plus the 200 validation molecules, and $\mathcal{N}=\mathcal{U}\setminus\mathcal{R}$ the novel molecules. Throughout, $|\cdot|$ denotes the number of items in a set, $A\setminus B$ the items in $A$ that are not in $B$, and $A\cap B$ the items in both. $\mathrm{Tc}(\cdot,\cdot)$ denotes Tanimoto similarity of ECFP4 fingerprints (radius 2, 2048 bits), and $\mathrm{nn}(g)=\max_{s\in\mathcal{T}}\mathrm{Tc}(g,s)$ the similarity of a generated molecule $g$ to its most similar training molecule $s$. The Tanimoto similarity threshold for genuine yield is $\texttt{thresh}=0.7$.}
\label{tab:metrics}
\small
\begin{tabular}{@{}p{0.26\linewidth}p{0.67\linewidth}@{}}
\toprule
Metric & Definition \\
\midrule

Validity & $|\mathcal{V}|/n$. Fraction of sampled strings that decode to a valid
molecule. \\[3pt]

Uniqueness & $|\mathcal{U}|/|\mathcal{V}|$. Fraction of valid molecules that are
distinct. \\[3pt]

Novelty & $|\mathcal{N}|/|\mathcal{U}|$. Fraction of distinct valid molecules absent from the reference set (train plus validation). \\[3pt]

Yield & $|\mathcal{N}| = n\ \times\text{Validity}\times\text{Uniqueness}\times\text{Novelty}$, reported per 10{,}000 samples. Count of novel, distinct, valid molecules that composes the three metrics above into one end-to-end measure. \\[3pt]

Genuine yield & $|\{g\in\mathcal{N}:\mathrm{nn}(g)<\texttt{thresh}\}|$ per 10{,}000 samples. It excludes novel molecules that are minor edits of a training molecule. \\[3pt]
Seed recovery & $|\mathcal{U}\cap\mathcal{T}|/|\mathcal{T}|$. Fraction of the 1{,}000 training molecules the model regenerates exactly. Near 1 indicates memorization. \\[3pt]

IntDiv$_{\text{mol}}$ & $1-\binom{m}{2}^{-1}\sum_{i<j}\mathrm{Tc}(g_i,g_j)$ over the $m=|\mathcal{N}|$ novel molecules $g_1,\dots,g_m$. Higher means more structurally diverse. \\[3pt]
IntDiv$_{\text{scaf}}$ & The same quantity over Bemis--Murcko scaffolds. Acyclic molecules having no scaffold are excluded. \\[3pt]
\bottomrule
\end{tabular}
\end{table}

\section{Results and Discussion}\label{sec:results}

We proceed in three steps: (1) we first examine whether representation-level randomization helps in the low-data regime; (2) we then probe which axis of randomization is responsible; and finally (3) we test whether the representation itself matters. Within each step we report significance tests over independently trained models and compare the physicochemical property distributions of the novel molecules against the training set.

\subsection{Augmentation with randomized SAFE improves generation in low-data regime }\label{subsec:1}
We first ask whether data augmentation via SAFE randomization improves generation when only 1,000 molecules are available for training. Holding the architecture, tokenizer scheme, and optimizer budget fixed, we trained three arms that differ only in their training corpus: the 1,000 canonical SAFE strings with no augmentation (\texttt{safe-1k}) and full-randomized SAFE corpora of $\simeq$25,000 (\texttt{safe-25k}) and $\simeq$50,000 (\texttt{safe-50k}) strings derived from the same 1,000 molecules. All three were trained for an identical number of optimizer steps, so the comparison is at matched compute rather than matched epochs. Each arm was trained five times from independent random initializations, and every trained model generated three independent sets of 10,000 molecules. Generation sets were averaged within a training run, so each reported value is a mean over five independently trained models. The results are summarized in Table~\ref{tab:scale}. 

\begin{table}[!htbp]
\centering
\caption{\textbf{Effect of corpus size on SAFE generation (augmentation sweep)}. Values are mean$\pm$SD over five independently trained models per arm and each model's value is the mean of three generation sets of 10,000 samples. Yield and genuine yield are counts per 10,000 samples while the remaining rows are fractions.}
\label{tab:scale}
\small
\begin{tabular*}{0.9\textwidth}{@{\extracolsep{\fill}}lccc}
\toprule
Arm & \texttt{safe-1k} & \texttt{safe-25k} & \texttt{safe-50k} \\
 & Canonical & Randomized & Randomized \\
Training corpus & (1{,}000) & ($\approx$25{,}000) & ($\approx$50{,}000) \\
\midrule
Validity & 0.9977$\pm${0.0002} & 0.9326$\pm${0.0025} & 0.9238$\pm${0.0029} \\
Uniqueness & 0.1022$\pm${0.0002} & 0.6257$\pm${0.0204} & 0.8649$\pm${0.0031} \\
Novelty & 0.0198$\pm${0.0021} & 0.8466$\pm${0.0069} & 0.9371$\pm${0.0007} \\
Yield & 20$\pm${2} & 4,941$\pm${189} & 7,488$\pm${31} \\
Genuine yield & 15$\pm${3} & 4,643$\pm${186} & 7,079$\pm${38} \\
Seed recovery & 0.9998$\pm${0.0003} & 0.8937$\pm${0.0143} & 0.5024$\pm${0.0044} \\
IntDiv$_{\text{mol}}$  & 0.9010$\pm${0.0044} & 0.8845$\pm${0.0006} & 0.8863$\pm${0.0003} \\
IntDiv$_{\text{scaf}}$ & 0.8789$\pm${0.0082} & 0.8632$\pm${0.0006} & 0.8655$\pm${0.0005} \\
\bottomrule
\end{tabular*}
\end{table}

Without augmentation, the model is effectively a lookup table. Validity is near-perfect (0.998), but only 10.2\% of the valid samples are distinct, and these $\approx$1,020 distinct molecules are almost exactly the 1,000 training molecules (seed recovery 0.9998) plus a handful of others. Just 2.0\% of these are novel for a yield of 20 molecules per 10,000 samples. High validity here reflects memorization rather than generative quality, which is why it must be read alongside uniqueness and novelty. Randomization changes this behavior. At $\approx$25,000 strings, uniqueness rises to 0.63 and novelty to 0.85, and at $\approx$50,000, to 0.86 and 0.94 while seed recovery falls to 0.89 and 0.50. Yield increases about 370-fold from \texttt{safe-1k} to \texttt{safe-50k}, and genuine yield, which excludes novel molecules within a Tanimoto similarity of 0.7 of a training molecule, increases about 458-fold. Internal diversity of the novel molecules is comparable between \texttt{safe-25k} and \texttt{safe-50k} at both the molecule level (0.884 vs.\ 0.886) and the scaffold level (0.863 vs.\ 0.866), so the additional yield at 50K comes from more novel molecules of similar diversity rather than from a narrower set. The close agreement between the two levels also indicates that this diversity is not merely substituent variation on a few ring systems. It establishes that SAFE augmentation shaped a memorizing model into a practical generator without narrowing or broadening the sampled chemical space. The model explores the same region of chemistry, but now produces new members of it rather than copies of its training set. The \texttt{safe-1k} diveristy values are not comparable since they are computed from only about 20 novel molecules per set. Validity declines modestly, from 0.998 to 0.933 and 0.924, as the model moves from reproducing its training set to producing new strings. This cost is small relative to the gains in uniqueness and novelty that yield captures.

\begin{figure}[!htbp]
\centering
\includegraphics[width=\textwidth]{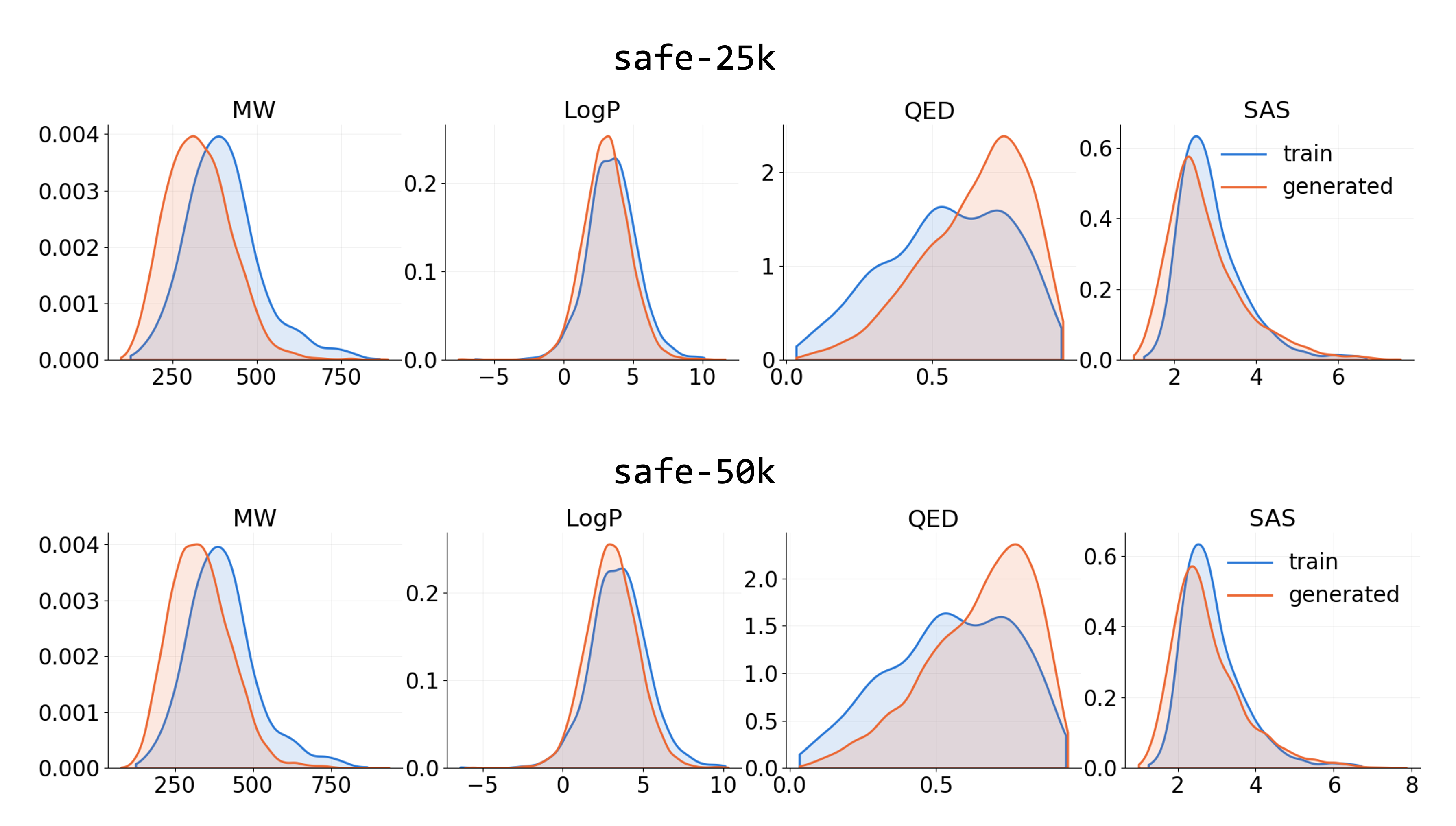}
\caption{\textbf{Physicochemical property distributions of the novel generated molecules}. Kernel density estimates of molecular weight, logP, QED, and synthetic accessibility score (SAS) for the novel molecules of \texttt{safe-25k} (top) and \texttt{safe-50k} (bottom) against the 1,000 training molecules. Each panel shows one representative generation set from a single trained model per arm. The \texttt{safe-1k} arm is omitted because it yields only about 20 novel molecules per generation set.}\label{fig:props-sweep}
\label{fig:props}
\end{figure}

These differences are found to be statistically significant. Treating the five independently trained models per arm as replicates, Welch's ANOVA with Games--Howell post-hoc comparisons separates all three corpus-size arms on yield ($F(2,5.4)=1.3\times10^{5}$, $p<10^{-12}$), genuine yield, uniqueness, novelty, and seed recovery, with every pairwise comparison resolved. The increments are large relative to their uncertainty. \texttt{safe-25k} exceeds \texttt{safe-1k} by 4,921 novel molecules per 10,000 samples (95\% CI 4,620--5,223), and \texttt{safe-50k} exceeds \texttt{safe-25k} by a further 2,547 (95\% CI 2,248--2,845). For genuine yield the corresponding increments are 4,628 (95\% CI 4,330--4,924) and 2,436 (95\% CI 2,144--2,729). Validity moves in the opposite direction and is also resolved, declining by 0.065 from \texttt{safe-1k} to \texttt{safe-25k} and by a further 0.009 to \texttt{safe-50k}. Because the replicates are independently trained models rather than repeated samples from a single model, these tests establish that the arms differ beyond both generation-set sampling noise and training-run variance. Run-to-run variation is itself informative. The standard deviation of yield across training runs is 31 molecules per 10,000 for \texttt{safe-50k} or 0.4\% of its mean, against 189 for \texttt{safe-25k}, or 3.8\% of its mean, so training-run variation is about an order of magnitude larger in relative terms for the intermediate corpus, where the model is between the memorizing and generating regimes. To check that the gain does not come at the cost of chemical plausibility, Figure~\ref{fig:props-sweep} compares the distributions of molecular weight, logP, QED, and synthetic accessibility for the novel molecules of \texttt{safe-25k} and \texttt{safe-50k} against the training molecules. \texttt{safe-1k} is omitted since it yields only about 20 novel molecules per set. The novel molecules shift toward lower molecular weight and correspondingly higher QED, while logP and synthetic accessibility track the training distribution closely, so the additional yield is drawn from the drug-like interior of the training distribution rather than from an unusual region of chemical space.

This low-data behavior stands in deliberate contrast to prior observations at benchmark scale. \citet{mesbahi_safe_2024} reported that SAFE randomization confers limited benefit when training on MOSES-scale corpora ($10^4-10^6$ molecules), attributing what benefit there is largely to extended training and reduced overfitting rather than to improved generalization. Our results are consistent with theirs rather than contradictory: \textit{the benefit of randomization is data regime-dependent}. When data is plentiful, a model sees enough distinct molecules that memorization of individual canonical strings is not the dominant failure mode, and randomization adds little. When only $\sim10^3$ molecules are available, a canonical-only model revisits each of its few strings many times and collapses onto them, and randomization is precisely what breaks this collapse. Representation-level augmentation, in other words, is most valuable exactly where generation is hardest: the low-data regime that characterizes most real lead-optimization and target-specific design settings.

\subsection{Joint atom-order and fragment-order randomization outperforms either axis alone}\label{subsec:axes}

We established that data augmentation via SAFE randomization improves molecular generation in the low-data regime. The question that now arises is which of the two axes for randomization (\texttt{dot} vs \texttt{atom}, that is, fragment-order vs atom-order) is primarily responsible for this improvement. To answer this, we trained three arms under the same five-seed protocol. Following the augmentation sweep, in which \texttt{safe-50k} performed best, we requested up to 50 variants per molecule on each axis and then matched the corpora molecule by molecule (Eqs.~\eqref{eq:kmin} and~\eqref{eq:nmatch}),
capping each molecule at the smallest count any axis can supply. This gives $N_{\mathrm{match}} = 32{,}152$ strings per arm, on average 32.2 variants per seed, so every arm sees the same number of strings for the same molecules and differs only in which axis those variants vary. Table~\ref{tab:axis} summarizes the results.

\begin{table}[!htbp]
\centering  
\caption{\textbf{Contribution of each randomization axis (axis ablation)}. All three arms are composition-matched to $N_{\mathrm{match}}=32{,}152$ strings per arm by Eqs.~\eqref{eq:kmin} and~\eqref{eq:nmatch}, so they differ only in which axis is randomized. Values are mean$\pm$SD over five independently trained models per arm and each model’s value is the mean of three generation sets of 10,000 samples.}
\label{tab:axis}
\small
\begin{tabular*}{0.9\textwidth}{@{\extracolsep{\fill}}lccc}
\toprule
Arm & \texttt{safe-atom} & \texttt{safe-dot} & \texttt{safe-full} \\
Randomized axis & Atom order & Fragment order & Both \\
\midrule
Validity                     & 0.9769$\pm$0.0017 & 0.9874$\pm$0.0006 & 0.9114$\pm$0.0031 \\
Uniqueness                   & 0.1974$\pm$0.0034 & 0.2171$\pm$0.0051 & 0.8674$\pm$0.0047 \\
Novelty                      & 0.5345$\pm$0.0071 & 0.5801$\pm$0.0087 & 0.9301$\pm$0.0021 \\
Yield & 1{,}031$\pm$30    & 1{,}244$\pm$47    & 7{,}353$\pm$43 \\
Genuine yield & 712$\pm$20        & 1{,}026$\pm$42    & 6{,}977$\pm$43 \\
Seed recovery                & 0.8973$\pm$0.0043 & 0.8997$\pm$0.0027 & 0.5525$\pm$0.0157 \\
IntDiv$_{\text{mol}}$        & 0.8872$\pm$0.0002 & 0.8843$\pm$0.0003 & 0.8842$\pm$0.0003 \\
IntDiv$_{\text{scaf}}$       & 0.8695$\pm$0.0010 & 0.8628$\pm$0.0013 & 0.8639$\pm$0.0005 \\
\bottomrule
\end{tabular*}
\vspace{1em}
\end{table}

The results show that neither axis of randomization alone is sufficient. Randomizing atom order within fixed, sorted fragment blocks gives a yield of 1,031 novel molecules per 10,000 samples and permuting fragment order while leaving each block canonically spelled gives 1,244. Randomizing both axes jointly gives 7,353, which is 7.1$\times$ the atom-order arm, 5.9$\times$ the fragment-order arm, and 3.2$\times$ the sum of the two. Each single axis falls short of the combination by more than 6,000 molecules per 10,000 samples (95\% CI 6,253--6,391 for atom order and 6,026--6,191 for fragment order). Genuine yield shows the same ordering more sharply (712, 1,026, and 6,977), which means that the gap is not an artifact of counting near-duplicates (or very chemically similar molecules) as novel.

The yield decomposition locates where the two single-axis arms fail. Relative to \texttt{safe-dot}, the full arm gains a factor of $4.0$ in uniqueness and $1.6$ in novelty while losing a factor of $0.92$ in validity, and these three together compose to the observed $5.9\times$. Uniqueness is therefore the binding constraint. With only one axis varied, roughly four out of five valid samples are repeats of a molecule the model has already emitted. Seed recovery tells the same story from the other side. Both single-axis arms regenerate about $90\%$ of the training set exactly ($0.897$ and $0.900$), close to the $0.9998$ of the unaugmented model in Table~\ref{tab:scale}, whereas the full arm recovers only $0.552$. Thirty-two variants per molecule along a single axis leave the model still largely reproducing its training set but the same budget spread across both axes does not. As in the augmentation sweep (Table~\ref{tab:scale}), the gain in yield is accompanied by a modest validity cost ($0.987$ to $0.911$) and leaves internal diversity essentially unchanged ($0.884$--$0.887$ at the molecule level, $0.863$-$0.870$ at the scaffold level), so the additional molecules are more numerous rather than more narrowly distributed.

\begin{figure}[!htbp]
\centering
\includegraphics[width=\textwidth]{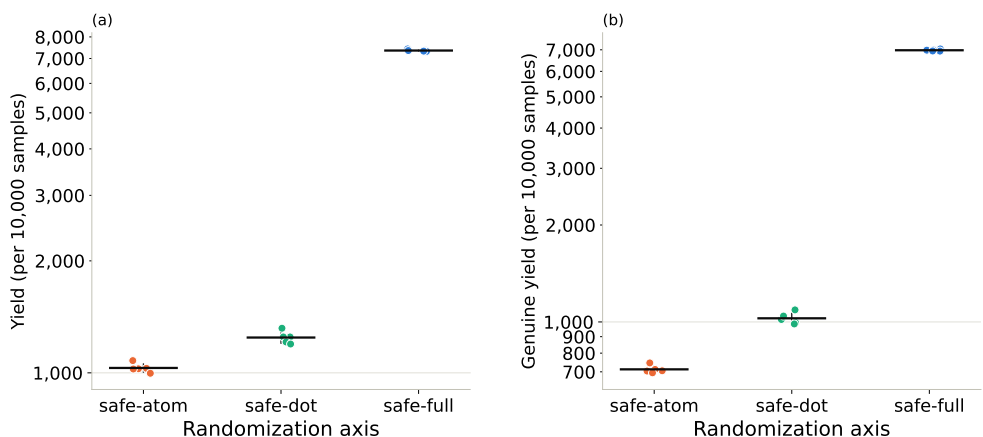}
\caption{\textbf{Per-run yield (left) and genuine yield (right) for the three axis-ablation arms}. Each point is one trained model, the mean of its three generation sets of 10,000 samples (five models per arm). Horizontal bars show the mean across models and whiskers show the standard deviations. The standard deviations (20 to 47 molecules per 10,000 samples) are small relative to the between-arm differences and are barely visible on the logarithmic $y$-axis.}
\label{fig:axis}
\end{figure}

Welch's ANOVA with Games--Howell post-hoc comparisons separates all three arms on yield ($F(2,7.6)=3.6\times10^{4}$, $p<10^{-15}$), genuine yield, uniqueness, and novelty. The fragment-order advantage over atom order is small but resolved, at 213 (1244 minus 1031) novel molecules per 10,000 samples (95\% CI 139--288) and 314 (1026 minus 712) for genuine yield (95\% CI 249--378). The advantage  of joint randomization over fragment order alone is an order of magnitude larger, at 6,109 (7,353 minus 1,244; 95\% CI 6,026--6,191). The one comparison that does not resolve is seed recovery between the two single-axis arms (\texttt{atom}--\texttt{dot} = -0.0024; 95\% CI -0.009--0.004), where both arms retain about 90\% of seeds, compared with 55\% under full randomization (differences of 0.345 and 0.347, 95\% CIs 0.320–0.369 and 0.322–0.372 respectively), consistent with both axes leaving memorization largely intact. Figure~\ref{fig:axis} shows the five per-run yield and genuine yield values for each arm, each the mean of three generation sets. The three arms form non-overlapping groups, with the full arm separated from both single-axis arms by roughly six to seven-fold.

\begin{figure}[!htbp]
\centering
\includegraphics[width=\textwidth]{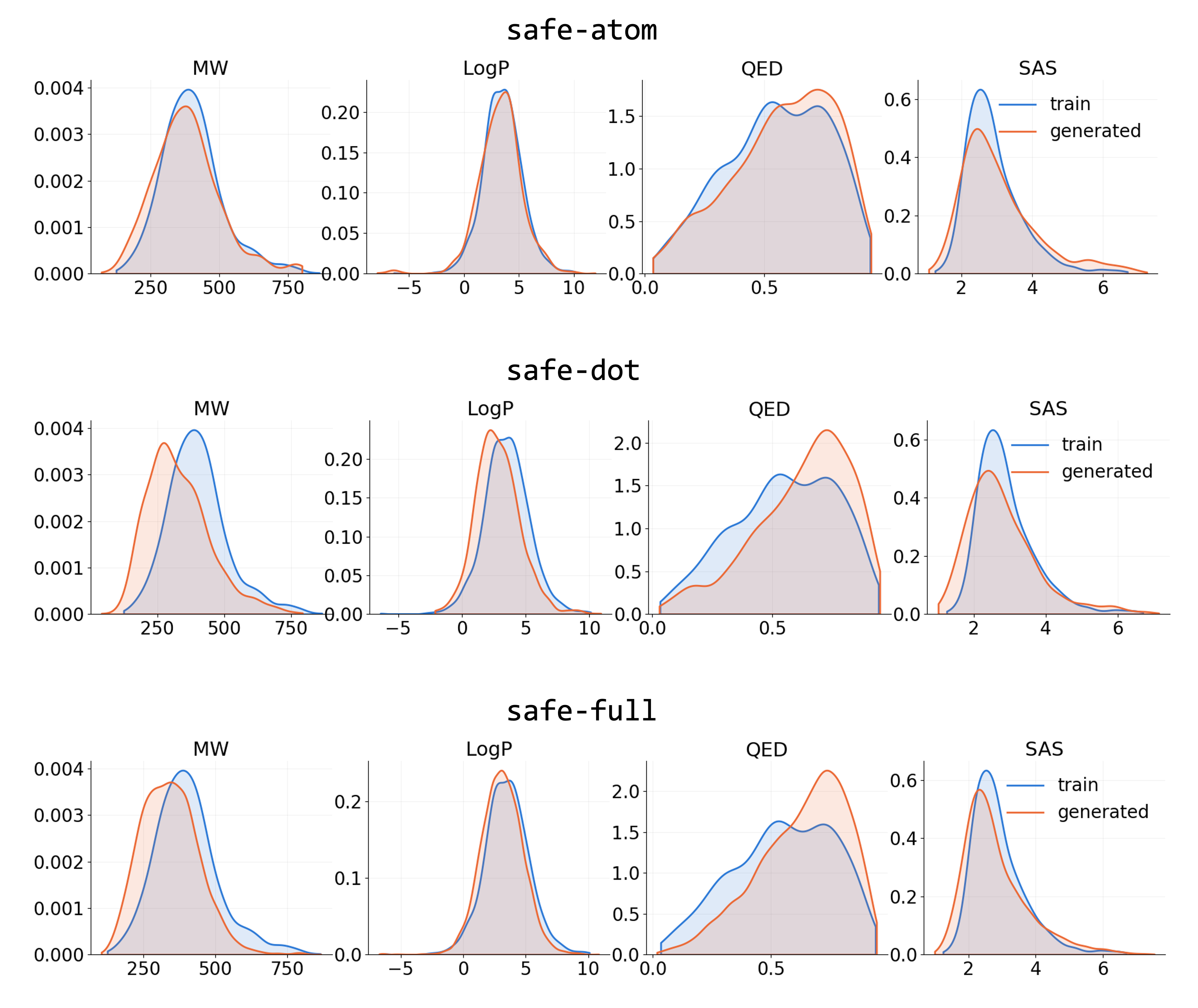}
\caption{\textbf{Physicochemical property distributions of the novel generated molecules}. Kernel density estimates of molecular weight, logP, QED, and synthetic accessibility score (SAS) for the novel molecules of \texttt{safe-atom} (top), \texttt{safe-dot} (middle), and \texttt{safe-full} (bottom) against the 1,000 training molecules. Each panel shows one representative generation set from a single trained model per arm.}
\label{fig:effort_2_props}
\end{figure}

Figure~\ref{fig:effort_2_props} compares the property distributions of the novel molecules from the three axis-ablation arms against the training set. All three remain within drug-like ranges in every property, so none of the randomization schemes drives the model into atypical chemistry. The arms differ mainly in molecular weight: \texttt{safe-atom} almost traces the training distribution, with a slightly lower peak and a broader spread, while \texttt{safe-dot} and \texttt{safe-full} shift toward lower values. QED moves in step with this, since QED penalizes high molecular weight, so the arms with the larger downward weight shift also carry more mass at high QED. LogP is essentially unchanged for \texttt{safe-atom} and \texttt{safe-full} and shifts slightly left for \texttt{safe-dot}, while synthetic accessibility is closest to the training distribution for \texttt{safe-full}. Each panel is drawn from a single generation set, so these differences are qualitative and the quantitative comparison is in Table~\ref{tab:axis}.

These results point towards an interaction between the two axes of SAFE randomization rather than an additive contribution from either. Because the per-molecule multiplicity is matched by construction, the arms differ not in how many strings the model sees per molecule but in how those strings are spread through the space of valid encodings. Atom-order variation respells fragments whose order is fixed, fragment-order variation reorders blocks whose spelling is fixed, and only their combination reaches encodings that differ in both respects at once. A formal two-by-two interaction test is not possible here, because the fourth cell representing no randomization on either axis cannot be matched to the same corpus size (a non-randomized SAFE corpus contains exactly one string per molecule and therefore cannot supply 32,152 strings from 1,000 seeds). The ablation is accordingly a comparison among the three realizable cells, and the super-additivity is descriptive. Even so, this is the central practical finding of this work: \textit{the fragment-order axis is what distinguishes SAFE from SMILES, but on its own it recovers only a sixth of the yield that joint randomization achieves, and an implementation that randomizes only fragment order would capture little of what the representation offers.}

\subsection{Randomized SAFE outperforms randomized SMILES by diversifying molecules rather than spellings}\label{sec:safe-vs-smiles}

Finally, we studied how data augmentation in SAFE (\texttt{safe-rand}) compares with data augmentation in SMILES (\texttt{smiles-rand}), shedding light on the comparative usefulness of the representation itself. For SAFE, we use \texttt{full} randomization mode
which randomizes both axes, because this configuration gave the best performance in the axis ablation (Section~\ref{subsec:axes}), exceeding either single axis. We compared the two schemes at a matched corpus size of 49,209 strings, again obtained by requesting up to 50 variants per molecule and matching the arms molecule by molecule, so any difference reflects what is randomized rather than how much training data each scheme happens to produce. Table~\ref{tab:smiles-v-safe} reports the metrics for both arms, each averaged over five independently trained models.

\begin{table}[!htbp]
\caption{\textbf{Randomized SMILES versus randomized SAFE at matched corpus size of 49,209 strings}. The arm \texttt{safe-rand} randomizes both axes (fragment order and atom order, i.e.\ \texttt{full}). Entries are mean$\pm$SD over five independently trained models per arm and each model's value is the mean of three generation sets of 10,000 samples.}
\label{tab:smiles-v-safe}
\small
\begin{tabular*}{0.9\textwidth}{@{\extracolsep{\fill}}lcc@{}}
\toprule
Arm                   & \texttt{smiles-rand} & \texttt{safe-rand} \\
Randomized axis       & Atom order           & Both               \\
\midrule
Validity              & 0.9484$\pm${0.0012} & 0.9230$\pm${0.0012} \\
Uniqueness            & 0.5096$\pm${0.0117} & 0.8650$\pm${0.0016} \\
Novelty               & 0.7993$\pm${0.0054} & 0.9365$\pm${0.0012} \\
Yield                 & 3,864$\pm${119}      & 7,477$\pm${24}       \\
Genuine yield         & 2,562$\pm${90}       & 7,067$\pm${32}       \\
Seed recovery         & 0.9697$\pm${0.0037} & 0.5067$\pm${0.0085} \\
IntDiv$_{\text{mol}}$     & 0.8814$\pm${0.0002} & 0.8868$\pm${0.0002} \\
IntDiv$_{\text{scaf}}$     & 0.8619$\pm${0.0002} & 0.8659$\pm${0.0005} \\
\botrule
\end{tabular*}
\end{table}

Randomized SAFE gives a yield of 7,477 novel molecules per 10,000 samples against 3,864 for randomized SMILES, a ratio of 1.94. The gap widens on genuine yield, where SAFE reaches 7,067 against 2,562, a ratio of 2.76. Because only two arms are compared, we use a two-sided Welch $t$-test on the per-run values, with no omnibus test or post-hoc adjustment. Both differences are large relative to the between-run spread (Welch $t(4.3) = 66.6$, $p = 1.1\times10^{-7}$ for yield and Welch $t(5.0) = 105.3$, $p = 1.7\times10^{-9}$ for genuine yield), corresponding to 3,613 more novel molecules per 10,000 samples (95\% CI 3,468--3,760) and 4,505 more genuine ones (95\% CI 4,396--4,616). The widening from 1.94 to 2.76 is the more informative of the two numbers. Because genuine yield discards novel molecules whose nearest training neighbor exceeds a Tanimoto similarity of 0.7, 34\% of the novel molecules from the SMILES arm are discarded against 5.5\% for SAFE. The advantage of fragment-level augmentation is therefore not only that more distinct valid strings are produced, but that a larger fraction of them sit outside the immediate neighborhood of the training molecules.

Randomized SMILES has the higher validity (0.9484 against 0.9230), so the SAFE advantage does not come from producing more valid strings. It comes from uniqueness (0.8650 against 0.5096) and novelty (0.9365 against 0.7993). Seed recovery shows this most directly. The SMILES arm regenerates 97.0\% of the seed molecules, against 50.7\% for SAFE, so a large part of its output reproduces the training set. Internal diversity over the novel molecules is 0.887 (molecule) and 0.866 (scaffold) for SAFE against 0.881 and 0.862 for SMILES, respectively. These values are descriptive rather than tested. They measure the spread within each set of novel molecules, not whether the two sets occupy the same region of chemical space. One might ask whether SAFE's larger vocabulary (51 against 38 for SMILES), rather than its grammar, accounts for its advantage. A smaller vocabulary makes next-token prediction easier, since the model chooses among fewer alternatives at each step and each token type recurs more often in a corpus of a given size, so the next-token distribution is both lower in entropy and better estimated. Therefore, any effect of vocabulary size would favor the SMILES arm, and the SMILES arm is in fact the one with the higher validity (0.9484 against 0.9230). The vocabulary difference therefore works against the result we report rather than producing it.

\begin{figure}[!htbp]
\centering
\includegraphics[width=\textwidth]{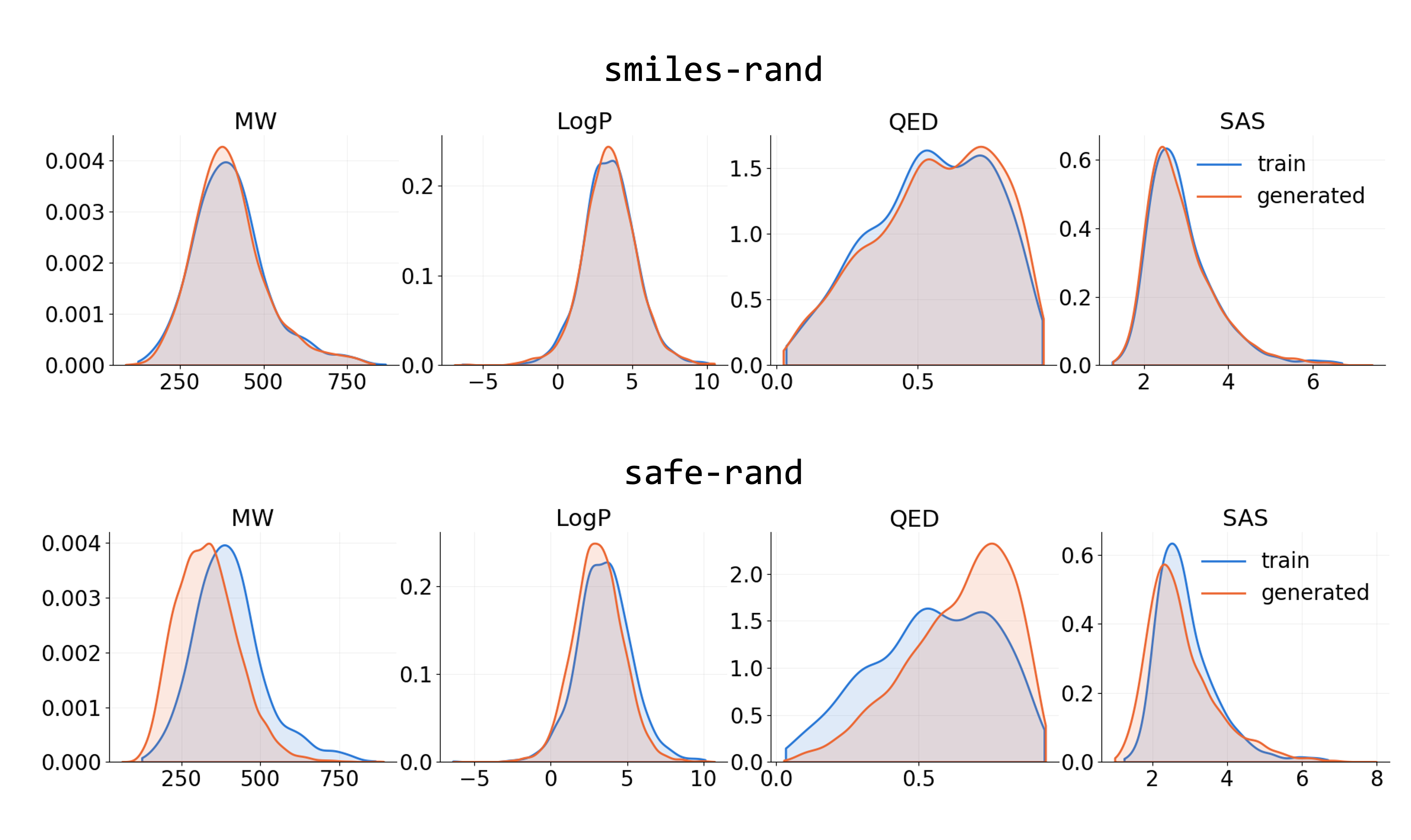}
\caption{\textbf{Physicochemical property distributions of the novel molecules at matched corpus size}. Kernel density estimates of molecular weight, logP, QED, and synthetic accessibility score (SAS) for the novel molecules of \texttt{smiles-rand} (top) and \texttt{safe-rand} (bottom), each plotted against the 1,000 training molecules. Each panel shows one representative generation set from a single trained model per arm.}
\label{fig:props-safe-smiles}
\end{figure}

Beyond the counts, the generated molecules can be examined directly. Figure~\ref{fig:props-safe-smiles} compares the property distributions of the novel molecules from \texttt{smiles-rand} and \texttt{safe-rand} against the training set. The two arms differ in how closely their novel molecules reproduce the training distribution. For \texttt{smiles-rand} the generated and training curves are nearly indistinguishable in all four properties, which is what one expects when 34\% of its novel molecules lie within a Tanimoto similarity of 0.7 of a seed and 97.0\% of the seeds are regenerated exactly. The property distribution is
largely inherited rather than learned. \texttt{safe-rand} departs modestly, toward lower molecular weight and correspondingly higher QED (QED penalizes high molecular weight), while logP and synthetic accessibility are essentially unchanged. The departure leaves the novel molecules well inside drug-like ranges in every property, so the larger genuine yield of the SAFE arm is not obtained by drifting into atypical chemistry.

\begin{figure}[!htbp]
\centering
\includegraphics[width=\textwidth]{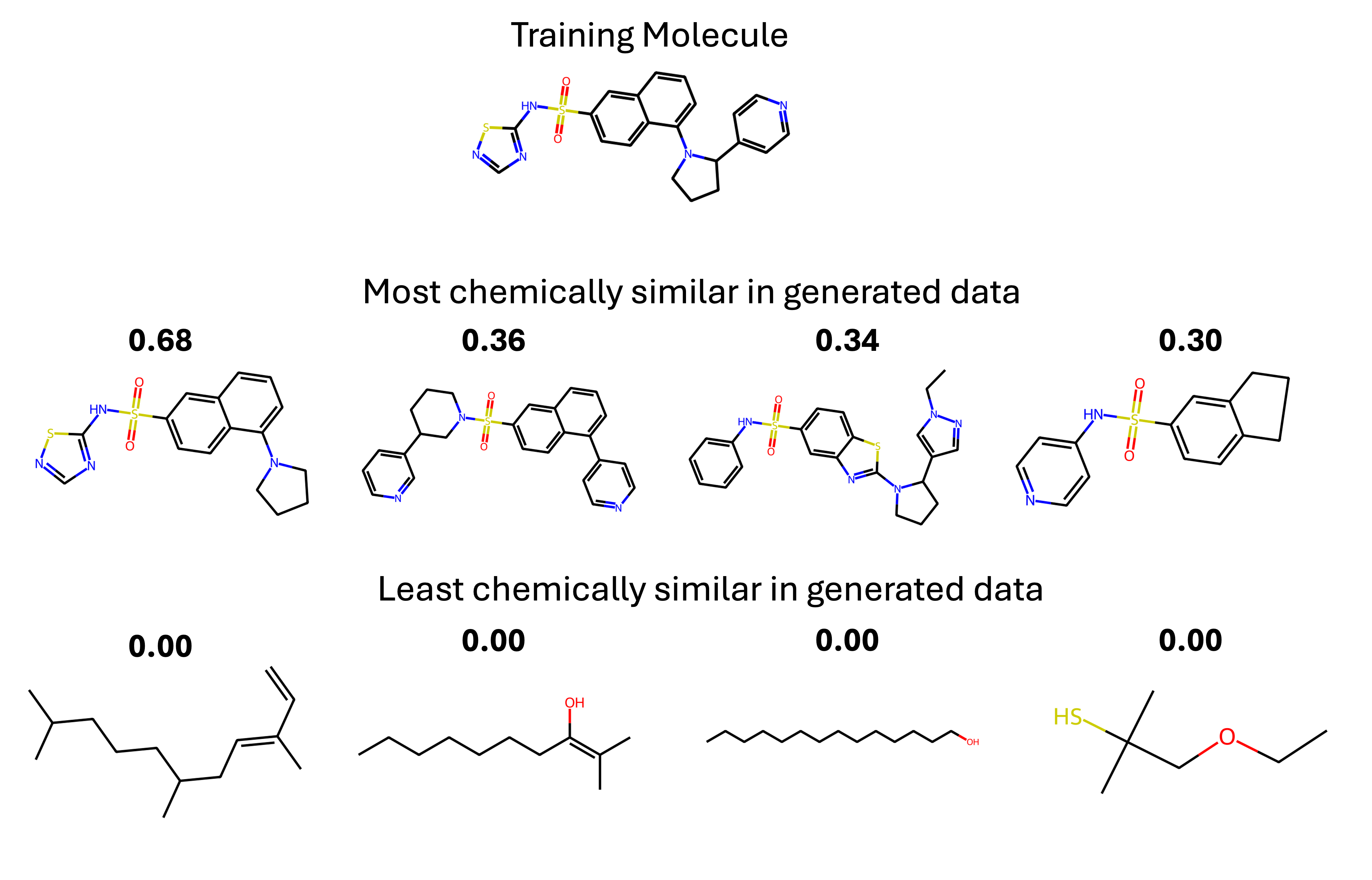}
\caption{\textbf{Generated molecules nearest to and furthest from a single, randomly sampled training molecule} (top). Middle row shows the four \texttt{safe-rand} molecules with the highest Tanimoto similarity to it and bottom row shows the four with the lowest. The value above each structure is its Tanimoto similarity (Morgan fingerprints, radius 2, 2048 bits) from the training molecule.}
\label{fig:neighbors}
\end{figure}

At the level of individual structures, Figure~\ref{fig:neighbors} takes a randomly sampled training molecule and shows the generated molecules nearest to it and furthest from it in Tanimoto similarity. The nearest ones retain the aryl sulfonamide linkage, with the closest also reproducing the naphthalene and thiadiazole of the query, while the fused ring system and the attached nitrogen heterocycle vary. Int his example, the closest generated molecule falls below the 0.7 near-duplicate threshold we have used in this work, and similarity drops to 0.36 by the second nearest. The furthest are acyclic aliphatic molecules sharing no ring system or sulfonamide with the query.

\begin{figure}[!htbp]
\centering
\includegraphics[width=\textwidth]{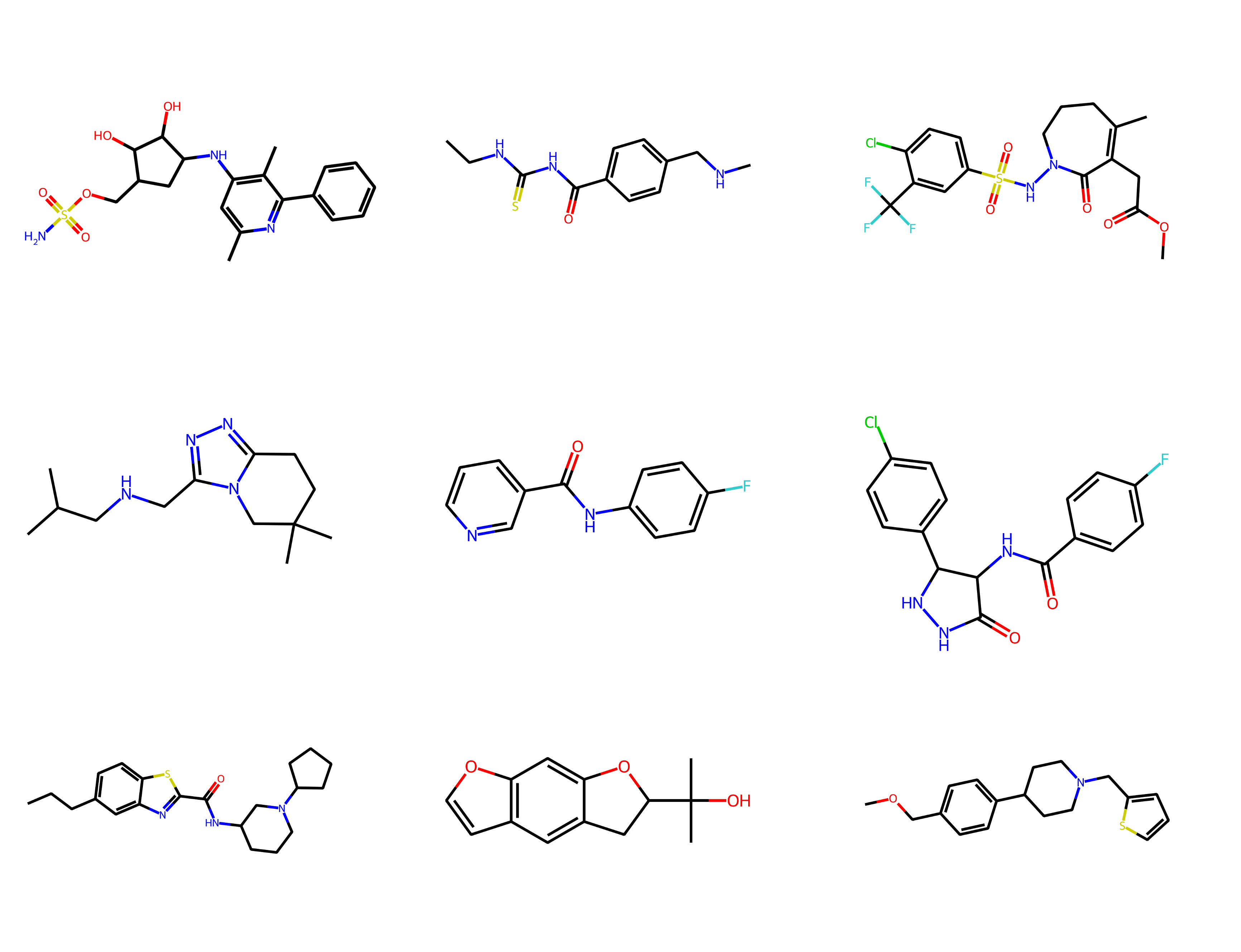}
\caption{\textbf{Sample novel molecules generated by \texttt{safe-rand}}. Nine molecules drawn uniformly at random from the generated set.}
\label{fig:novel}
\end{figure}

Figure~\ref{fig:novel} shows a sample of novel molecules from the same arm, spanning fused heteroaromatic systems, saturated nitrogen heterocycles, sulfonamides and amides, and halogenated aromatics. These two figures are illustrative and are not evidence of a distributional property, which is what Table~\ref{tab:smiles-v-safe} and Figure~\ref{fig:props-safe-smiles} show.

\section{Conclusions}\label{sec:conclusion}

We asked three targeted questions in this work and addressed them. Working from 1,000 seed molecules, with the architecture, tokenization, hyperparameters, and training budget held constant, and with every arm trained five times from independent initializations, we asked:
(1) \emph{Does randomized SAFE augmentation help a decoder-only chemical language model in a strictly low-data regime?} It does, and decisively. A 50$\times$ expansion of the corpus by data randomization alone, without a single additional molecule, converts a model that is effectively a lookup table into a working generator, raising the yield of genuinely novel molecules by 458 fold.
(2) \emph{Which of its two randomization axes is responsible?} Neither alone.
With corpora matched molecule by molecule, varying both axes gave a genuine yield 6.8$\times$ that of fragment-order randomization, 9.8$\times$ that of atom-order randomization, and 4$\times$ their sum.
(3) \emph{Does the SAFE representation itself matter once augmentation is held fixed?} It does. At a matched corpus size, randomized SAFE reaches 2.8$\times$ the genuine yield of randomized SMILES.

Together, these results identify the design of augmentation and not simply the number of augmented strings, as an important variable in low-data molecular generation. Two results carry the argument, both resting on matched training corpora. The first is the large gap between joint SAFE randomization and either single-axis condition. The second is the advantage of randomized SAFE over randomized SMILES at matched corpus size, which is larger on genuine yield than on yield and therefore isolates what fragment-order randomization contributes. Genuine yield discards novel molecules that are near-copies of a training seed molecule, and it discards a far larger share of the SMILES arm's output than of SAFE's. Randomizing atom order presents the model with many spellings of the same molecule and it learns the spellings. Randomizing fragment order additionally presents many orderings of the parts, and it learns to recombine them. This is also why the two axes are not additive. The reading rests on the matched design: with every arm seeing the same number of strings for the same molecules, nothing remains to explain the gap except what those strings vary. That the benefit appears here and not at benchmark scale~\cite{mesbahi_safe_2024} is a statement about data regime rather than a contradiction, since randomization pays where a model would otherwise memorize, and on the few hundred to few thousand molecules a real campaign supplies, it would.

Several limitations bound these conclusions. First, the arms in the representation comparison are matched in strings rather than tokens, and because SAFE strings are about 22\% longer than the corresponding randomized SMILES (Table~\ref{tab:token-lengths}), the SAFE arm processes correspondingly more tokens at an identical number of optimizer steps. We therefore cannot separate the contribution of the representation from that of the larger token budget it entails. This overhead is intrinsic to the SAFE encoding rather than an experimental choice, so matching the arms on tokens would require unmatching their molecular composition, which will not be justified. Second, the axis ablation compares the three realizable cells of a two-by-two design, since a corpus randomized on neither axis contains exactly one string per molecule and cannot be matched to the same number of strings, so the super-additivity we report is descriptive rather than the outcome of a formal interaction test. Third, all results use a single seed set drawn from GuacaMol, a single fragmentation scheme, one model size, and unconditional generation, and genuine yield depends on a Tanimoto threshold of 0.7 that we did not sweep (Section~\ref{subsec:eval}). Because each of these choices is held identical across every arm, they bound how far the conclusions generalize but cannot account for the differences between arms, which is what the matched design is
constructed to isolate. Finally, BRICS in particular fixes the fragment granularity and therefore bounds the diversity that fragment-order randomization can express. A coarser or finer slicing scheme would change the size of the effect, though we see no reason it would change its direction.

Within these bounds, the practical recommendation is clear. An implementation that randomizes only fragment order, which is the axis unique to SAFE and the one most readily associated with it, recovers about a sixth of the yield that joint randomization achieves, and one that randomizes only atom order recovers even less. The benefit of SAFE in this regime is realized only when both axes vary together. The limitations above also set the agenda. Whether the same augmentation benefits encoder-only SAFE-based models~\cite{kaushal_fragberta_2026}, how the effect decays as the seed set grows from $10^3$ toward the benchmark scale at which its benefit has been reported to be limited~\cite{mesbahi_safe_2024}, whether it survives when generation is conditioned on a scaffold or a property target, and how it responds to coarser or finer fragmentation are all open and all testable within the same matched-composition framework, and is the subject of our future work. The broader point this study makes is methodological --- \textit{string augmentation is not a single knob but a choice of what to vary, and in the data-poor settings where generative models are most needed, that choice determines whether augmentation produces new spellings or new molecules.}

\backmatter

\bmhead{Supplementary information}
The Supplementary Information contains the full statistical tables for the three families/studies, and the training configuration and sweep design (model, optimization, sampling and replication settings for all 40 training runs).

\section*{Declarations}

\bmhead{Availability of Data and Materials}
All data, code and results supporting this article are publicly available in the \texttt{safe-lowdata-augmentation} repository (\url{https://github.com/kaushal-mind-lab/safe-lowdata-augmentation}). The repository contains the seed and validation molecule sets, the scripts that construct all eight training corpora, the tokenizers, the pipeline for training, generation and evaluation, the metrics for all 120 generation sets and 40 training runs, and the statistical analysis code and outputs.

\bmhead{Competing interests}
The authors declare no competing interests.

\bmhead{Authors' Contributions}
N.K. conceived and designed the study, developed the methodology, performed the experiments, and wrote the original draft of the manuscript. A.T. and A.P. designed and conducted the statistical analysis, and postprocessing of results including data visualization and properties extraction. All authors contributed to, reviewed, edited, and approved the final version.

\bmhead{Acknowledgements}
This work was supported in part by the NVIDIA Academic Grant Program, which provided 32,000 A100 GPU-hours on the NVIDIA Brev platform. It also used the Delta advanced computing and data resource which is supported by the National Science Foundation (award OAC 2005572) and the State of Illinois. Delta is a joint effort of the University of Illinois Urbana-Champaign and its National Center for Supercomputing Applications. Access to Delta was provided through the National Science Foundation-funded ACCESS program (Project MED250048).

\bmhead{Use of Generative AI}
The authors used Claude (Anthropic) to assist with code development and to improve the readability of the manuscript, and ChatGPT (OpenAI) for language editing. The authors reviewed and edited the content as needed and take full responsibility for the content of the publication.


\bibliography{references_fragopt}

\end{document}